\documentclass[]{style/ceurart}
\usepackage{listings}
\usepackage{graphicx}
\usepackage{bbm}

\begin{document}

\copyrightyear{2024}
\copyrightclause{Copyright for this paper by its authors.
  Use permitted under Creative Commons License Attribution 4.0
  International (CC BY 4.0).}

\conference{CLEF 2024: Conference and Labs of the Evaluation Forum, September 9-12, 2024, Grenoble, France}

\title{Tile Compression and Embeddings for Multi-Label Classification in GeoLifeCLEF 2024}

\author[1]{Anthony Miyaguchi}[
orcid=0000-0002-9165-8718,
email=acmiyaguchi@gatech.edu,
]
\cormark[1]
\author[1]{Patcharapong Aphiwetsa}[
email=paphiwetsa3@gatech.edu,
]
\author[1]{Mark McDuffie}[
email=mmcduffie8@gatech.edu,
]

\address[1]{Georgia Institute of Technology, North Ave NW, Atlanta, GA 30332}
\cortext[1]{Corresponding author.}

\begin{abstract}
We explore methods to solve the multi-label classification task posed by the GeoLifeCLEF 2024 competition with the DS@GT team, which aims to predict the presence and absence of plant species at specific locations using spatial and temporal remote sensing data. 
Our approach uses frequency-domain coefficients via the Discrete Cosine Transform (DCT) to compress and pre-compute the raw input data for convolutional neural networks.
We also investigate nearest neighborhood models via locality-sensitive hashing (LSH) for prediction and to aid in the self-supervised contrastive learning of embeddings through tile2vec.
Our best competition model utilized geolocation features with a leaderboard score of 0.152 and a best post-competition score of 0.161.
Source code and models are available at \url{https://github.com/dsgt-kaggle-clef/geolifeclef-2024}.
\end{abstract}


\maketitle

\section{Introduction}

GeoLifeCLEF \cite{geolifeclef2024} is a task organized within the LifeCLEF lab \cite{lifeclef2024} at the CLEF 2024 conference, with the objective of predicting which plant species are present and absent in specific locations given spatial and temporal remote sensing data. 
Modeling species density distributions can be helpful in biodiversity management and conservation.

We explore methods to solve the posed multi-label classification task and incorporate unsupervised methods to build useful representations from the data. 
We propose a pipeline that pre-computes tiles from raw GeoTIFF images and stores a compressed version on disk to speed up the training process.
We utilize metadata to build baseline geolocation models and indices for nearest-neighbor queries.
Our models utilize convolutional neural networks to exploit spatial information, spectral representations, and co-located bands of remote sensing data.
We also explore the use of unsupervised methods to learn representations for knowledge transfer between two different datasets with similar semantics.

\section{Related Works}

GeoLifeCLEF 2023 had seven submissions along with baseline results from the organizers \cite{geolifeclef2023}. 
Participants focused on bioclimatic rasters and satellite imagery, leveraging Convolutional Neural Networks (CNN) like ResNets \cite{he2016deep} for feature extraction. 
Participants combined rasters and trained separate models for prediction \cite{Ung2023LeverageSW}. 
Spatial coordinates (longitude/latitude) were commonly used with models like K-Nearest Neighbors (kNN) and Random Forest, yielding surprisingly good results. 
However, the combination of diverse modalities provided in the dataset was rare, with only one participant utilizing time-series data with a 1D Convolutional Network.

\section{Overview}

\begin{figure}[h]
\centering
  \begin{minipage}{0.49\linewidth}
    \begin{lstlisting}[frame=single]

{
  "type": "Polygon",
  "coordinates": [
      [
          [-32.26344, 26.63842],
          [-32.26344, 72.18392],
          [ 35.58677, 72.18392],
          [ 35.58677, 26.63842],
          [-32.26344, 26.63842],
      ]
  ],
}

    \end{lstlisting}
    \caption{GeoJSON polygon definition.}
    \label{lst:geojson-definition}
  \end{minipage}
  \hfill
  \begin{minipage}{0.49\linewidth}
      \centering
      \includegraphics[width=1\linewidth]{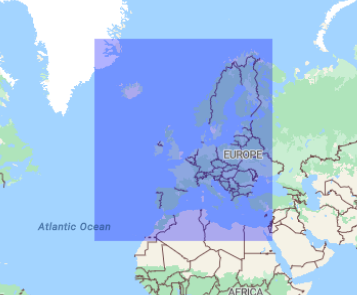}
      \caption{Polygon region overlayed on a map.}
      \label{fig:geojson-region}
  \end{minipage}
\end{figure}

The competition has three main components to the dataset. 
The first are the metadata associated with the competition comprising a presence-only training, presence-absence training, and presence-absence test set. 
The metadata provides a mapping between location and the species labels available for supervised training. 
The second are the remote-sensing and raster data provided in pixel format. 
The final component is time series data containing quarterly environmental data over a 20 year period. 

The presence-only training dataset comprises 5,079,797 examples over 3,845,533 survey sites distributed throughout Western Europe.
The dataset is drawn from crowd-sourced data with potential gaps in the reported species.
The presence-absent dataset has stricter semantics -- species not included in the survey are presumed absent.
The training set has 1,483,637 examples over 88,987 sites, while the test set has 4,716 sites.
The datasets include an identifier for the survey site along with latitude and longitude, according to the schema in Listing \ref{lst:schema}.
We compute a projection into EPSG 3035, which allows for Euclidean distance between sites in units of meters. 

\begin{figure}[b]
\begin{lstlisting}[caption={
  Metadata schema for the competition.
}, captionpos=b,label={lst:schema},frame=single]
  |-- dataset: string (nullable = true)
  |-- surveyId: integer (nullable = true)
  |-- lat_proj: double (nullable = true)
  |-- lon_proj: double (nullable = true)
  |-- lat: double (nullable = true)
  |-- lon: double (nullable = true)
  |-- year: integer (nullable = true)
  |-- geoUncertaintyInM: double (nullable = true)
  |-- speciesId: double (nullable = true) 
\end{lstlisting}
\end{figure}

The majority of available data are raster and satellite imagery. 
The GeoTIFF files provided contain various measures such as elevation, roads, population, and soil. 
The GeoTIFF files are bounded by a GeoJSON polygon that covers Western Europe as seen in Figures \ref{lst:geojson-definition} and \ref{fig:geojson-region}.
RGB-NIR satellite imagery is directly available as $128\times128$ pixel tiles associated with each survey site.

\section{Processing Pipeline}


\begin{figure}[h!]
  \centering
  \title{Modeling Pipeline}
  \includegraphics[width=\textwidth]{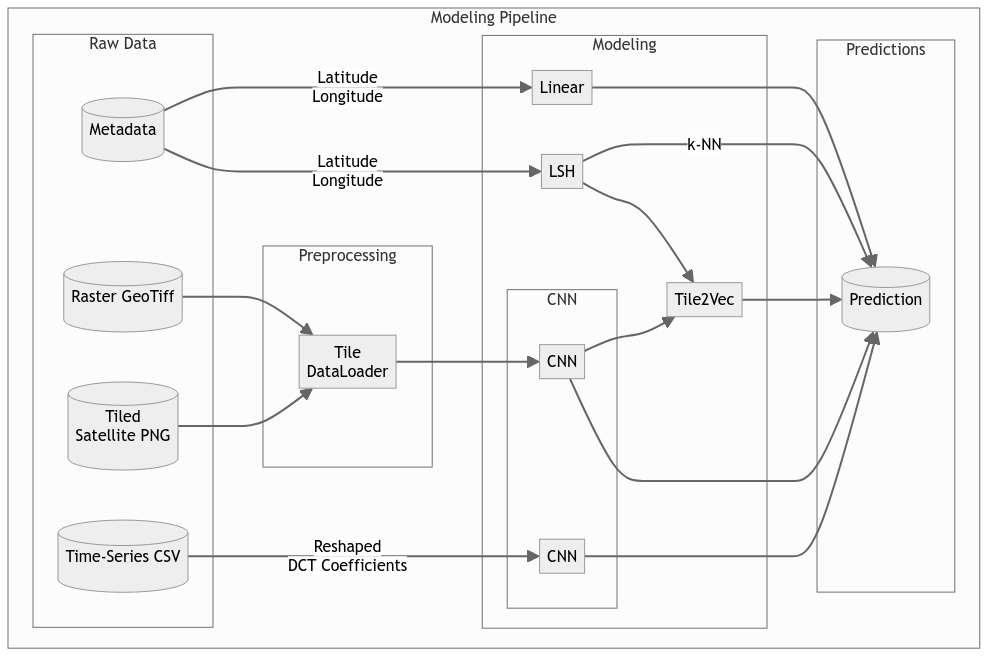}
  \caption{
    Overview of the data and modeling pipeline.
    Raw data is pre-processed to maintain survey site per row semantics, with 2D DCT coefficients as features.
    Data is cached as columnar parquet files in cloud storage for efficient access.
  }
  \label{fig:tiled-raster}
\end{figure}

We explore several solutions for the multi-label classification problem. 
We use Luigi \cite{Rouhani2024spotify} as our workflow management tool, which provides idempotent directed acyclic graphs (DAGs) of tasks. 
We use Spark \cite{armbrust2015spark} to perform data extraction, transformation, and loading (ETL) from tarred images and CSV files to columnar parquet files. 
We use PyTorch as our deep learning framework and use PyTorch-Lightning to simplify the training and inference process.
We use Petastorm to preprocess and load data into Torch.
We use Weights and Biases to log hyperparameters and metrics.

\subsection{Satellite and Raster Data}

\begin{figure}[h!]
  \centering
  \includegraphics[width=\textwidth]{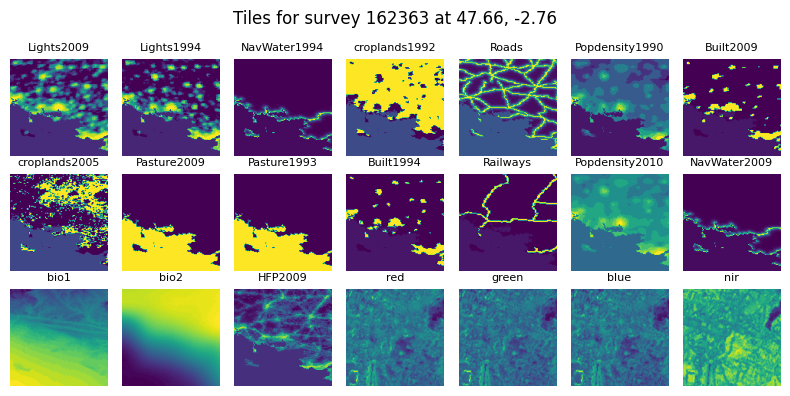}
  \caption{
      Example of a tiled raster image. 
      The image is a 128x128 tile of the RGB-NIR satellite imagery. 
      The image is associated with a survey site and is used as input to the model. 
  }
  \label{fig:tiled-raster}
\end{figure}

The competition organizers provide point data for each survey site in a pre-computed train CSV file.
Our experiments focus on 128x128 pixel tiles extracted from provided GeoTIFF files for use in a supervised learning setting.
Significant in-memory overhead tiling exists because we need to store a 128x128 matrix of integers or floats for each survey site.
Memory access patterns can cause significant slow-downs if we need to fetch them from disks often. 

We fork the official \texttt{plantnet/GeoLifeCLEF} data loaders to pre-compute tiles for each of the provided GeoTIFF under 1GB. 
Certain rasters do not fit into memory (e.g., elevation raster at 11GB); therefore, we omit them from our experiments. 
We only compute the tiles associated with the survey identifiers in our metadata, which helps limit the size of the resulting dataset. 

\subsection{Tile Compression via Discrete Cosine Transform (DCT)}

\begin{figure}[h!]
  \centering
  \includegraphics[width=0.3\textwidth]{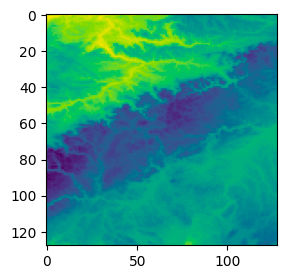}
  \hfill
  \includegraphics[width=0.3\textwidth]{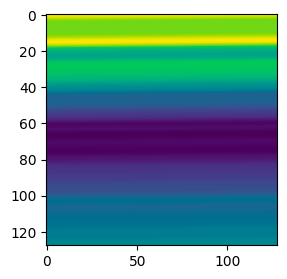}
  \hfill
  \includegraphics[width=0.3\textwidth]{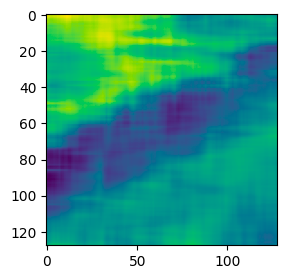}
  \caption{
    Example of low-pass filtering using the DCT.
    (a) Original bio1 raster image.
    (b) Low-pass filter using the first 50 coefficients of the 1D-DCT, reshaping on the first axis (row-major order).
    (c) Low-pass filter using the 2D-DCT using the top-left 8x8 coefficients.
  }
  \label{fig:dct-lowpass}
\end{figure}

We computed the 2D-DCT on the resulting tile images and kept low-frequency coefficients as features in downstream modeling. 
We implement a PySpark wrapper around the ND-DCT to supplement the standard library 1D-DCT implementation for feature pre-processing.
We lose significant spatial information if we perform filtering in 1D coefficient-space as seen in Figure \ref{fig:dct-lowpass}.

\subsection{Time-Series Data}

\begin{figure}[h!]
  \centering
  \includegraphics[width=0.495\textwidth]{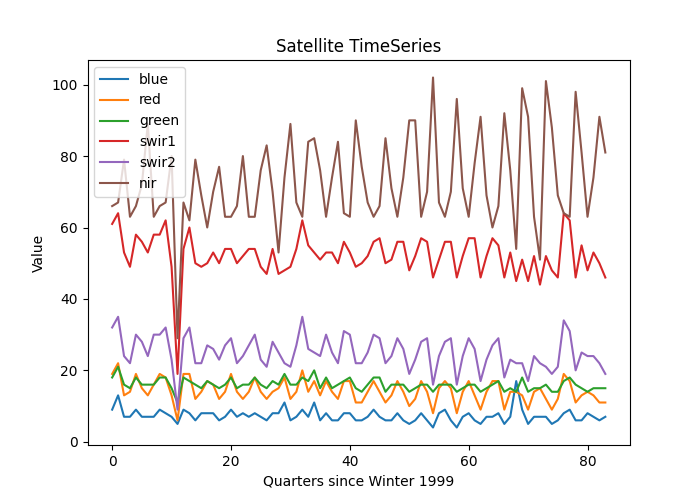}
  \hfill
  \includegraphics[width=0.495\textwidth]{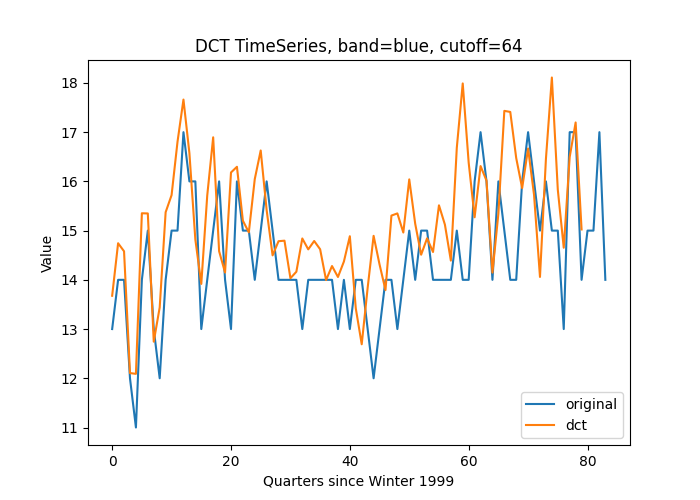}
  \caption{
    Example of time-series data and its DCT.
    (a) Original time-series data.
    (b) Blue band time-series approximated using DCT.
  }
  \label{fig:dct-timeseries}
\end{figure}

Time series data are treated as another layer in the network by pre-processing the data to obtain DCT coefficients.
We have access to quarterly time series data for each survey site over 20 years.
Some sites have missing data, which are padded with zeros.
We compute the 1D DCT on the time series data and keep the first 64 coefficients in the transformed space, which parallels the 8x8 2D-DCT coefficients extracted from the raster data.
The original time-series data and its DCT are displayed side by side in figure \ref{fig:dct-timeseries}

\subsection{Data Augmentation}

\begin{figure}[h!]
\begin{lstlisting}[language=Python,frame=single]
class DCTHorizontalFlip:
    def __init__(self, k=8):
        self.odd_factor = -torch.ones((k, k))
        for i in range(0, k, 2):
            self.odd_factor[i, :] = 1

    def forward(self, X):
        return X * self.odd_factor
\end{lstlisting}
\caption{Augmentation of 2D-DCT coefficients that flips the image across the horizontal axis in the spatial domain.}
\label{lst:augmentation}
\end{figure}

We apply augmentations to our data to encourage model invariance to rotation and reflection.
Some such transforms include rotating and flipping images before sending them into a model.
Equivalent augmentations exist in the frequency space.
For example, a 90-degree rotation in pixel space is equivalent to the transpose of the 2D-DCT coefficients.
We can flip the image in pixel space by alternating the signs of the 2D-DCT coefficients along a given axis.
Rotations and flips along the axis give us enough flexibility to implement useful symmetries in the data to improve model generalization.

\subsection{Locality-Sensitive Hashing for Nearest Neighbor Queries}

Species at sites close together should intuitively have similar distributions of plants. 
The projected latitude and longitude have physical meaning via euclidean distance, so we can build a nearest-neighbor model and rank species per survey site by frequency in a neighborhood.
We use locality-sensitive hashing (LSH) with random hyperplane projections to build a k-NN model \cite{leskovec2020mining}, with the hyper-parameters for bucket length and number of hash tables set to 20 and 5, respectively.
We can find the top-k nearest survey sites in linear time for any given survey site using the LSH model.
The approximate nearest neighbor self-join is performed with a 50km cutoff, and the results stored on disk for downstream modeling.

\section{Models}

\subsection{Nearest Neighbor Model}

We generate predictions by querying the LSH model built on survey site locations across presence-only and presence-absent datasets in a 50km radius.
For each survey site, we limit either the number of neighbors or the distance to the nearest neighbor.
Our nearest neighbor models (NN) consider all neighbors within a 5km, 10km, and 50km radius.
The k nearest neighbors model (k-NN) considers the top 10 neighbors for each survey site and all of the species reported in the neighbors.

\subsection{Geolocation Model}

We model the relationship between geospatial metadata (i.e. projected latitude and longitude) and the species labels.
We learn a linear and non-linear multi-label classification problem, and treat this model as a baseline for remote-sensing based models.
The linear model uses a single linear layer that maps the feature space into the output space.
The nonlinear model uses a two-layer model, with a $\mathcal{R}^{256}$ latent space in the first layer and a linear layer to map to the output space.
We add random noise to latitude and longitude with a mean of 5km to increase generalization.

\subsubsection{Tile CNN Model}
\label{sec:tile-cnn}

The processed remote-sensing data are represented by a multi-dimensional array with one indexing dimension for the layer type and two spatial dimensions, making this a natural fit for CNNs.
We pre-process satellite and raster data by generating 128x128 pixel tiles, then applying the 2D-DCT to filter the 8x8 coefficients representing the lowest spatial frequencies.
We use the coefficients as inputs to a CNN model that learns a mapping to a multi-label classification task.
We additionally run a few experiments where we apply the inverse 2D-DCT to the coefficients to recover a multi-layer 128x128 pixel image.
We learn models on the presence-absent dataset, since the dataset is representative of the test.

We build CNN models that convolve input layers with a 3x3 kernel with padding, followed by a 1x1 convolution and linear layer to map into latent space.
We map the latent space to a linear layer with the number of classes as the output space.
Batch normalization is applied at every convolution for numerical stability and ReLU activation for nonlinearities.
We also experiment with alternative parameterizations of the network, notably replacing the custom CNN with a pre-packaged efficientnetv2 backbone.

We experiment with several models that take in different input data.
The simplest model uses RGB-NIR satellite imagery in four channels.
We then incorporate the 13 MODIS landcover layers and 19 bio-climatic rasters.
We then hand-choose specific layers from the rasters to remove redundancy, specifically layers 9, 10, and 11 from MODIS and the years 2001, 2010, and 2019 from the bio-climatic rasters.
The other layers of the MODIS dataset correspond to legacy classification schemes and confidence bands, which are likely not useful to our model.
We choose three bands from bio-climatic rasters since the years are evenly spaced across the set.

We built a model using the provided time-series RGB-SWIR data.
We reused the same architecture as the satellite imagery data by reshaping the first 64 coefficients of the time-series DCT into an 8x8 tensor and then applied the same convolutional layers.
The semantics are not necessarily the same as the 2D-DCT coefficients, but we hypothesize learned structure from this basis despite the lack of spatial symmetry.

\subsubsection{Tile2Vec Model}

Tile2Vec \cite{jean2019tile2vec} is a self-supervised learning technique that learns embeddings of tiles of satellite imagery.
The Tile2Vec model utilizes a spatially-aware sampling procedure and triplet loss to learn a low-dimensional embedding that preserves metric distances via the triangle inequality.
The triplet loss is $L(t_a, t_n, t_d)$ with a margin $m$ where $f_{\theta}$ maps data to a $d$-dimensional vector of real numbers using a model with parameters $\theta$. 

\begin{equation}
    L(t_a, t_n, t_d) = \left[
        ||f_{\theta}(t_a) - f_{\theta}(t_n)||_2
        - ||f_{\theta}(t_a) - f_{\theta}(t_d)||_2
        +m
    \right]_+
\end{equation}

We obtain triplets from the presence-only datasets by querying the LSH model to sample one million pairs of tiles within 100km in the presence-only dataset.
We generate a distant neighbor by randomly selecting a tile from dataloader batch.
We train a tile2vec model using the CNN architecture described in section \ref{sec:tile-cnn} without a classifier head in $\mathcal{R}^{256}$ latent-space.
We experiment with a multiobjective loss incorporating the sum of triplet and ASL losses, using labels for each survey site by aggregating all species within each site's radius.
The classifier adds a linear layer to the learned latent space and training with the ASL loss on the presence-absent dataset.

\subsection{Model Evaluation and Loss Functions}

The competition uses the F1-micro score to evaluate models, and we use the same metric in training.
We evaluated the model during training and validation using \texttt{MulticlassF1Score(average="micro")} from the \texttt{torchmetrics} library, with a 90-10 train-validation split of the presence-absent dataset.

\subsubsection{Binary Cross-Entropy}

Binary cross-entropy is a loss function used for binary classification.
In a multi-label setting, each label as an independent binary classification problem.
We use the loss function that accepts logits as input for numerical stability, which is necessary to achieve acceptable convergence.

\begin{equation}
L = -\sum_{c=1}^My_{o,c}\log(p_{o,c})
\end{equation}

\subsubsection{Asymmetric Loss (ASL)}

The asymmetric loss \cite{ridnik2021asymmetric} penalizes false positives and false negatives differently from the binary cross-entropy loss.
The loss is defined in terms of the probability of the network output $p$, and hyperparameters $\gamma_{+}$ and $\gamma_{-}$.
Setting $\gamma_{+} > \gamma_{-}$ emphasizes positive examples, while setting both terms to 0 yields binary cross-entropy.
Easy negative samples are dynamically down-weighted, and hard thresholded samples are ignored.

\begin{equation}
  ASL=\left\{
    \begin{array}{l}
      L_{+}=(1-p)^{\gamma_{+}} \log (p) \\
      L_{-}=\left(p_m\right)^{\gamma_{-}} \log \left(1-p_m\right)
    \end{array}\right.
\end{equation}

We sweep over parameters $\gamma_{+} \in \{0, 1\}$ and $\gamma_{-} \in \{0, 2, 4\}$.
The default values are $\gamma_{+}=1$ and $\gamma_{-}=4$.

\subsection{Hill Loss}

The Hill loss \cite{zhang2021simple} is a loss function designed for robust multi-label classification with missing labels.
The loss is defined as a weighted mean-squared error (MSE), where the weight modulates potential false negatives.

\begin{equation}
    \begin{aligned}
    \mathcal{L}_{\text {Hill }}^{-} & =-w(p) \times M S E \\
    & =-(\lambda-p) p^2 .
    \end{aligned}
\end{equation}

The implementation provided by the authors provides the following form with default values of $\lambda=1.5$ and $\gamma=2$.

\begin{equation}
    \mathcal{L}_{\text {Hill }}^{-} = y \times (1-p_{m})^\gamma\log(p_{m}) + (1-y) \times -(\lambda-p){p}^2
\end{equation}

\subsubsection{sigmoidF1}

The sigmoidF1 loss \cite{benedict2021sigmoidf1} optimizes the F1 score directly by creating a differentiable approximation of the F1 score.
We first define the terms true positive, false positive, false negative, and true negative as a function of the sigmoid function.

\begin{equation}
  \begin{aligned}
  & \widetilde{t p}=\sum \mathbf{S}(\hat{\mathbf{y}}) \odot \mathbf{y} \quad
  \tilde{f p}=\sum \mathbf{S}(\hat{\mathbf{y}}) \odot(\mathbbm{1}-\mathbf{y})
  \\
  & \tilde{f n} =\sum(\mathbbm{1}-\mathbf{S}(\hat{\mathbf{y}})) \odot \mathbf{y} \quad
  \tilde{t n}=\sum(\mathbbm{1}-\mathbf{S}(\hat{\mathbf{y}})) \odot(\mathbbm{1}-\mathbf{y})
  \end{aligned}
\end{equation}

where $\mathbf{S}(\hat{\mathbf{y}})$ is the sigmoid function applied to the model's output $\hat{\mathbf{y}}$.

\begin{equation}
  S(u ; \beta, \eta)=\frac{1}{1+\exp (-\beta(u+\eta))}
\end{equation}

Then we define the F1 score as a function of the true positive, false positive, and false negative terms.

\begin{equation}
  \mathcal{L}_{\widetilde{F 1}}=1-\widetilde{F 1}, \quad \text { where } \quad \widetilde{F 1}=\frac{2 \widetilde{t p}}{2 \widetilde{t p}+\widetilde{f n}+\widetilde{f p}}
\end{equation}

We are given two hyper-parameter $S=-\beta$ and $E=\eta$.
For tuning, we sweep over parameters $S \in \{-1, -15, -30\}$ and $E \in \{0, 1\}$ as suggested in the author's experiments.
The default values are $S=-1$ and $E=0$.


\section{Results}

We report the performance of our models using the hidden test set on the Kaggle competition leaderboard.
For torch-based models, predictions are made in two ways: top-k and threshold.
The top-k method selects the top-k species with the highest probability, while the threshold method selects species with a probability greater than a threshold.
We set k to 20 and the threshold to 0.5 for all relevant models.

\subsection{Nearest Neighbor Model}

We report nearest neighbor models in table \ref{tab:nn}.
The k-NN models perform better than our NN models, probably due to filtering out noise from different thresholds.
We observe that when we do not limit by the number of neighbors, larger distance thresholds lead to worse performance, with a difference of 0.10 to 0.08 when going from 5km to 50km.
The opposite is true once we keep the top 10 neighbors, where there is a small improvement in score as we increase the distance threshold.

\begin{table}[h]
    \caption{
        Nearest neighbor Kaggle leaderboard scores using the LSH model on survey site distances in kilometers.
        We use cutoffs of 5km, 10km, and 50km to limit the number of neighbors.
        The k-NN model limits the number of neighbors to the top 10.
    }
    \label{tab:nn}
    \begin{tabular}{|l|l|l|}
    \hline
    \textbf{model} & \textbf{private} & \textbf{public} \\ \hline
    NN 50km         & 0.08638          & 0.08785         \\ \hline
    NN 10km         & 0.10719          & 0.10149         \\ \hline
    NN 5km          & 0.11059          & 0.10718         \\ \hline
    k-NN 50km        & 0.12919          & 0.1251          \\ \hline
    k-NN 10km        & 0.12545          & 0.11998         \\ \hline
    k-NN 5km         & 0.12219          & 0.11746         \\ \hline
    \end{tabular}%
\end{table}

\subsection{Geolocation Model}

The projected latitude and longitude provide a relatively strong signal for the multi-label species classification tasks, with a score of 0.161 on the public leaderboard when using the top-k results in table \ref{tab:cnn}.
Our experiments show that the linear model performs poorly as per table \ref{tab:losses}.
However, adding a non-linear layer to the model increases the performance by a large margin.
For example, models trained with BCE loss without class weights go from 0.03 to 0.14 when adding a nonlinear layer.

During training of BCE models, we observed that while the loss was decreasing between training and validation sets, the validation F1 score would peak on the first epoch and then decrease over time.
We suspect that BCE loss has difficulty with the class imbalance even when explicit class weights are given.
ASL increases the validation scores by a wide margin, likely because of the dynamic weighting behavior.
Although the Hill and sigmoidF1 losses report improvements over ASL in the experimental setting in the literature, we find that default hyperparameters perform worse than BCE loss.
It is possible hyper-parameter tuning could increase performance over ASL.
For the rest of the experiments, we focus on ASL as the primary loss due to its performance and robust parameterization.

\begin{table}[h]
    \caption{
        Scores for different losses on geolocation based models.
        The scores provided are collected from late submissions on the public leaderboard using the threshold method of prediction.
    }
    \label{tab:losses}
    \centering
    \begin{tabular}{|l|l|l|}
    \hline
    \textbf{Loss}               & \textbf{linear} & \textbf{nonlinear} \\ \hline
    BCE (weights= True)             & 0.0259          & 0.14523            \\ \hline
    BCE (weights=False)              & 0.03011         & 0.14398            \\ \hline
    ASL ($\gamma_{-}=4, \gamma_{+}=1$) & 0.00042         & 0.15768            \\ \hline
    ASL ($\gamma_{-}=4, \gamma_{+}=0$) & 0.00169         & 0.15619            \\ \hline
    ASL ($\gamma_{-}=2, \gamma_{+}=1$) & 0.00064         & 0.14963            \\ \hline
    ASL ($\gamma_{-}=2, \gamma_{+}=0$) & 0.0008          & 0.15414            \\ \hline
    ASL ($\gamma_{-}=0, \gamma_{+}=1$) & 0.00635         & 0.14873            \\ \hline
    ASL ($\gamma_{-}=0, \gamma_{+}=0$) & 0.00029         & 0.15793            \\ \hline
    Hill ($\lambda=1.5$)                  & 0.0048          & 0.14867            \\ \hline
    sigmoidF1 (E=1 S=-30)          & 0.00046         & 0.0014             \\ \hline
    sigmoidF1 (E=1 S=-15)          & 0.00404         & 0.00106            \\ \hline
    sigmoidF1 (E=1 S=-1)          & 0.00015         & 0.05031            \\ \hline
    sigmoidF1 (E=0 S=-30)          & 0.00146         & 0.03082            \\ \hline
    sigmoidF1 (E=0 S=-15)          & 0.00047         & 0.03917            \\ \hline
    sigmoidF1 (E=0 S=-1)          & 0.00177         & 0.0439             \\ \hline
    \end{tabular}
\end{table}

For BCE models, we find that adding a class weight that is proportional to the normalized frequency in the dataset does not improve performance of the model.
It's possible that these weights are not computed correctly, but ASL provides a dynamic weighting mechanism that is far more effective when the number of classes is large.

For our ASL models, we find that the best hyper-parameters are $\gamma_{-}=0$ and $\gamma_{+}=0$ on the public leaderboard, but the default value of $\gamma_{-}=4$ and $\gamma_{+}=1$ works just as well.
This score performs better than the BCE model, despite claims that setting values of $\gamma_{-}$ and $\gamma_{+}$ to zero would lead to a model that is equivalent to the BCE model.
We note that if we had to choose hyperparameters for ASL through a validation set, it's possible that we could choose one that would be sub-optimal for the test set.
We choose to use the default values for the rest of our experimentation.

The Hill loss performs between BCE and ASL in the non-linear model, and so we do not consider it further given the performance of ASL.
The sigmoidF1 loss performs the worst out of all of the losses, despite the tuning in the ranges provided by the literature.

\begin{table}[h]
    \caption{
        Kaggle private and public leaderboard scores for torch-based models.
        The models are trained on the presence-absent dataset.
    }
    \label{tab:cnn}
    \centering
    \begin{tabular}{|l|l|l|}
    \hline
    \textbf{model}     & \textbf{private} & \textbf{public} \\ \hline
    Geolocation ASL Top-k       & 0.16047          & 0.16128         \\ \hline
    Geolocation ASL Threshold     & 0.13395          & 0.13734         \\ \hline
    RGB-IR Top-k     & 0.16229          & 0.16108         \\ \hline
    RGB-IR Threshold   & 0.15217          & 0.15096         \\ \hline
    RGB-IR Landcover Top-k   & 0.01986          & 0.02096         \\ \hline
    RGB-IR Landcover Threshold & 0.02545          & 0.0255          \\ \hline
    Time-series Top-k            & 0.10497          & 0.10392         \\ \hline
    Time-series Threshold          & 0.08444          & 0.0832          \\ \hline
    Tile2Vec RGB-IR Top-k      & 0.14071          & 0.13943         \\ \hline
    Tile2Vec RGB-IR Threshold    & 0.12989          & 0.12701         \\ \hline
    Tile2Vec RGB-IR Top-k      & 0.14448          & 0.13832         \\ \hline
    Tile2Vec RGB-IR Threshold    & 0.13318          & 0.12438         \\ \hline
    \end{tabular}%
    \end{table}

\subsection{Tile CNN Models}

We report minor improvements in the performance of the geolocation model, with our best model utilizing satellite imagery at 0.161 on the public leaderboard in table \ref{tab:cnn}.
However, training a model on the presence-absent dataset is difficult due to model convergence to a minima.
This behavior is most prevalent in the efficientnetv2 backbone, where the larger parameter space and domain-specific preprocessing distortion lead to suboptimal convergence.

The first 13 channels of the landcover raster to the RGB-NIR channels does not converge to a useful model.
We find that our performance drops down to 0.02 when we try to utilize these features.
Keeping the subset of features from landcover aids in a model that performs relatively well but lacks large improvements over the RGB-NIR model.

The time series model learns some structure despite the strange input representation with a score of 0.10 on the public leaderboard.

\subsection{Tile2Vec Model}

Tile2vec learns a useful representation that leads to smooth convergence of downstream classifiers.
We observe convergence occurring within four epochs in our experimental setting, and the increase in the F1 metric for both validation and training sets increases monotonically on the classifier.
In contrast, models without tile2vec backbone have validation F1 scores that fluctuate, typically within the first five epochs, and then decrease over time.
The learned predictions are marginally less effective than learning the CNN model directly on the presence-absent dataset.

When we trained the model with ASL as part of the optimization objective, we observed that the triplet loss term no longer decreased monotonically over time.
Instead, it sharply decreased to a minimum, increased, and decreased slowly over time.
The behavior is likely due to the difference in magnitude of the triplet loss and the ASL loss.
The triplet loss is normalized, while the ASL loss is not, so the ASL hyperparameter dominates the gradient updates.
This version of the model performs better on the transfer learning task to present-absent data than the triplet loss alone.

\subsection{Competition Performance}

Our best models are reported against the public leaderboard in table \ref{tab:leaderboard}.
The best score of the competition is 0.4089, while baseline models provided by the competition organizers lies around 0.25.
Our models lie between the granular and coarse-grain frequency-based models.

\begin{table}[h]
    \caption{
        Kaggle private and public leaderboard scores with best models compared to baselines and top teams.
        Models included in the results are post-competition submissions.
    }
    \label{tab:leaderboard}
    \begin{tabular}{|l|l|}
    \hline
    \textbf{Name}                                             & \textbf{Public} \\ \hline
    Rank 1 - webmaking                                        & 0.4089          \\ \hline
    Rank 2 - AI2Lab                                           & 0.36837         \\ \hline
    Baseline with Landsat Cubes                               & 0.26576         \\ \hline
    Baseline with Bioclimatic Cubes                           & 0.2594          \\ \hline
    Baseline with Sentinel Images                             & 0.23629         \\ \hline
    Top-25 species per district \& biogeographical zones (PA) & 0.20302         \\ \hline
    Top-25 species per Country/Region (PA)                    & 0.17514         \\ \hline
    (Ours) Geolocation ASL Topk                                              & 0.16128         \\ \hline
    (Ours) RGB-IR Top-k                                             & 0.16108         \\ \hline
    (Ours) Tile2Vec RGB-IR Top-k                                             & 0.13943         \\ \hline
    (Ours) k-NN 50k                                                   & 0.1251          \\ \hline
    Top-25 species in Presence Absence                        & 0.11614         \\ \hline
    Top-25 species in Presence Only                           & 0.08133         \\ \hline
    \end{tabular}%
\end{table}

\section{Discussion}

We find difficulty overcoming basic baselines in the competition.
In particular, frequency-based baseline submissions can be significantly more effective than the solutions proposed in our research of the problem.
These solutions are done by predicting the top 25 species at varying levels of locality (e.g., globally or regionally) and by dataset.
However, we find that latitude and longitude are surprisingly predictive of plant species in the dataset given an appropriate loss function.
Using these geospatial features provides a useful diagnostic for more complex datasets, since the number of input features are small and are easier to debug.
One possible limitation of our methodology is that we do not utilize the presence-only dataset with the exception of pre-training the tile2vec model.

\subsection{Alternatives Methods}

Learning a relationship between latitude and longitude to species labels with classical machine learning techniques and standard libraries is computationally intractable.
We note alternative approaches that were explored but did not produce results for various reasons.

\subsubsection{Classical Supervised Learning}

Intead of using a neural network to learn a mapping from location to species, we tried learning the mapping via logistic regression.
This numerically simple model can be learned using Spark via stochastic gradient descent (SGD). 
As a validation, we build a model to predict the 10 most frequent species in the dataset per site using only the location features. 
This achieves an F1-macro score of 0.09 when splitting the sites into a 90-10 train-validation split, which is better than random but roughly equivalent to always choosing the most frequent species.

We run into out-of-memory (OOM) issues when learning 5 million rows and 10,358 species with scikit-learn or statsmodels.
When we run the same procedure in Spark via distributed stochastic gradient descent (SGD), we find it will run for over 48 hours on a GCP n1-standard-8 instance (8 vCPU, 16GB RAM, 350GB NVME SSD) using 3-fold cross-validation (CV). 
We suspect that this is due to the size of the coefficients that involve J features and K output classes. 
Presuming an 8-byte double, the coefficients alone will be at least 8MB, larger than the typical CPU cache. 

We investigate other algorithms for modeling multilabel classification, including Naive Bayes, SVM, Random Forests, and Factorization Machines.
Naive Bayes assumes non-negative count data. 
SVMs are not tractable for our problem and are slower to solve than linear/logistic regression for other problems in the Spark toolbox. 
Random forests only support up to 100 classes in Spark, likely due to the branching factor to support each class. 
Factorization machines suffer from a similar issue as logistic regression and SVMs. 
Our final attempt to model multi-class classification via classical supervised techniques is through XGBoost \cite{chen2016xgboost}, which maintains a Spark binding. 
We run out of memory when trying to model many classes.

\subsubsection{Low-rank Multilabel-Space Regression}

Instead of trying to learn the mapping between features and label-space directly, it is possible to learn a relationship between features and a low-rank multi-label space instead \cite{dasgupta2023review}.
It takes 30 minutes to learn a regression between location and a single binary response using either linear regression or XGBoost.
Given these constraints, we would like to constrain our model to 4-8 response variables. 

We try reducing the label space via the DCT since the relationship is trivially invertible in the machine learning pipeline. 
We find that this is untenable since we need many more coefficients than are available in our budget to represent discontinuities in species presence.


Another approach is to use singular value decomposition (SVD) to compute a projection of label-space into the first few eigenvectors, and then learn the relationship between the features and the projection \cite{tai2012multilabel}.
Then, predictions are quantized using nearest neighbors in the projection of the label space.
This process is similar to latent semantic indexing (LSI), and would allow a model to take into consideration cooccurrences between labels. 
While interesting, this approach requires significant engineering effort for results that are no more interpretable than neural networks.

\subsubsection{Node2Vec}

Node2vec \cite{grover2016node2vec} learns to preserve properties of network nodes using biased random walks.
Using the K-NN graph, we attempt embedding the survey sites using the co-occurrences of species among sites.
We could then use the embedding of the survey site as a feature for the classification task.
The embedding of the survey node is intractable due to the size of the network of 4 million nodes and ~1 billion edges. 
A species node embedding can be computed in 20 minutes, which results in a vector representation of species that can be used for clustering or classification.

A survey node embedding would be useful as a feature of the classification task since it would require no further processing to go from the survey site to the species. 
To take advantage of the species embeddings, we would need to compute some average of the embedding vectors before passing into a supervised classification model.

\section{Future Work}

We have explored various techniques for finding useful representations to model species distribution.
One area for future work is to capture better nuance associated with the self-supervised representation learning of the tiles.
We quickly reached a limit in how well the model could represent our training data, so it would be helpful to rigorously explore alternative model parameterizations and hyperparameters for the various loss functions.
Additionally, it is unclear how best to incorporate the many raster layers provided in the competition.
Ideally we would be able to determine which layers are most important to the multi-label classification task, possibly through extensive ablation testing of the features.

We would also like to continue down network or graph models of the survey and species.
A rich interconnection exists between sites and species where sparse co-occurrences could be exploited through spatial locality.
One way this could be done is by constructing node features through message passing of survey site nearest neighbors.
Graph neural networks could also be an effective mechanism for generating embedding spaces by propagating information through diffusion.
However, implementing these techniques could be challenging, especially since we failed to generate a survey embedding through the survey-species bipartite network due to computational constraints.

Our findings indicate a significant variation in the occurrence of the labels, with some labels with less than 10 data points and others with more than 10,000. 
Thus, this causes bias and imbalance in the training.
A possible solution would be to bin the labels according to their frequency so that each label is relatively in the same range in terms of data points. 
This would also allow us to utilize XGBoost since it would reduce the number of classes that need to be classified.

It would also be interesting to implement a proper model of the dynamics of the remote sensing data.
We can build manifold representation of satellite imagery, demonstrated by our experiments with tile2vec.
It should be possible to model the linearized dynamics of a system by learning a Koopman operator that steps forward state space from one timestep to another.
We hypothesize that this could be done by conditioning the tile embeddings on state evolutions, e.g., the 20 years of bioclimatic rasters and quarterly time series data.
One potential way to do this is to learn a spatio-temporal embedding of the tiles via an explicit sequence model like an LSTM or transformer alongside methods to enforce the geographical distributional semantics afforded by tile embeddings.
Another approach is to perform data-driven system identification to understand the dynamics of bioclimatic rasters that have been embedded in the space and to understand the governing equations of the system with a method like SINDy \cite{brunton2016discovering}.
\section{Conclusions}

In this study, we addressed the multi-label classification challenge of GeoLifeCLEF 2024, which aims to predict the presence or absence of plant species at specific locations based on spatial and temporal remote sensing data. 
We explored using a compressed version of the remote sensing data to train deep learning models, with varying levels of success.
We take advantage of the geospatial nature of the data by building a neighborhood model with locality-sensitive hashing.
Predictions from the neighborhood model perform better than some of the simplest frequency models made by the competition organizers.
The neighborhood model is used as part of a self-supervised embedding model that learns a low-dimensional representation of the data that is effective for classification.
Despite poor performance on the leaderboard, some of the ideas presented in this working note have potential for future work and have not been fully explored.
Source code and models are available at \url{https://github.com/dsgt-kaggle-clef/geolifeclef-2024}.

\section*{Acknowledgements}

Thank you to Professor Patricio Vela for supervising the project for Anthony Miyaguchi's ECE8903 Special Problems course at Georgia Tech.
Thank you to the DS@GT CLEF group for access to cloud computing resources through Google Cloud Platform, and for a supportive environment for collaboration.

\bibliography{main}


\end{document}